\documentclass[11pt]{article}
\usepackage{acl2015}
\usepackage{times}
\usepackage{url}
\usepackage{latexsym}
\usepackage{graphicx}
\usepackage{subcaption}
\usepackage[font=small,labelfont=bf]{caption}
\usepackage{enumitem}

\title{\LARGE TransCouplet: Transformer based Chinese Couplet Generation}

\author{
  Kuan-Yu Chiang, Shihao Lin, Joe Chen, Qian Yin, Qizhen Jin\\
  Dept. of Computer Science, University of Southern California \\
  \texttt{kuanyu@usc.edu}, \texttt{shihaol@usc.edu}, \texttt{zchen462@usc.edu}\\ \texttt{qianyin@usc.edu},\texttt{qizhenji@usc.edu} \\
  
}
 
\date{2021/12/02}

\begin{document}
\maketitle

\begin{abstract}
 Chinese couplet is a special form of poetry composed of complex syntax with ancient Chinese language. Due to the complexity of semantic and grammatical rules, creation of a suitable couplet is a formidable challenge. This paper presents an transformer-based sequence-to-sequence automatic couplet generation model. With the utilization of PinYin and Part-of-Speech tagging, the model achieves the couplet generation. Moreover, we evaluate the AnchiBERT \footnote{https://github.com/huggingface/transformers} on the ancient Chinese language understanding to further improve the model. 

\end{abstract}

\section{Introduction}

Among the diversity of literature, the Chinese couplet stands out with a special poetry form using pairs of orderly sentences. With the nurture of over 1000 years, the Chinese couplet has become a unique art and historical treasure, however, due to the complexity of semantic and grammatical rules, the creation of a suitable couplet is a formidable challenge. A couplet is an ancient Chinese literature form composed by a pair of two consistent lines. The first line is called Upper Line and the second is called Lower Line in a couplet. Sometimes the lines can be short sentences, but most of the time, they are just in the form of phrases or poetry lines. As a form of literature, a couplet has a nature of subjectivity, judging a match of a couplet is hard. With such a unique and symmetric structure, we are very interested in exploring how we can use text generation techniques we learned in this class to create a system that automatically generates the second line of a couplet given the first one.

\begin{figure}[htp]
    \includegraphics[width=7cm]{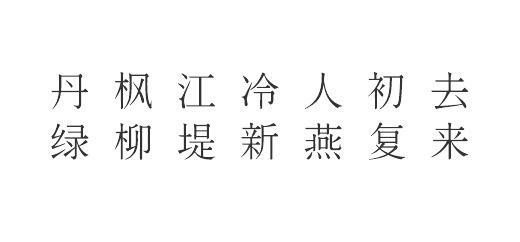}
    \caption{\small An example of Chinese Couplet. The meaning of the upper sentence is "In the red maple forest by the river, people began to leave for the cold weather.", followed by "On the embankment planted with green willows, swallows return when spring is coming."}
    \label{fig:sample}
\end{figure}

The main goal of this project is to explore and compare the performance of different systems on couplet text generation, and add on our own ideas and techniques to improve those systems. The two lines in a couplet are strictly symmetric in their structures: they must have the same amount of words (Chinese characters), grammar, as well as rhythm. More interestingly, two lines can be similar with or contradicted to each other in terms of the meanings. With such symmetric nature, we look at couplet generation as a Seq2Seq problem in NLP fields where we have two sequences with exactly the same number of words.

All code can be found on github \footnote{https://github.com/chiang9/NLP-Chinese\char`_couplet\char`_generation}, and the introduction video is also uploaded. \footnote{https://www.youtube.com/watch?v=GxvntrldVWg}

\section{Related Work}

In the last decade, with the availability of large collections of datasets, there have been several different approaches to the generation of Chinese couplets that have had success. 

Based on the Statistical Machine Translation model, Zhou, M. et al.~\shortcite{Zhou:09} used a phrase-based SMT to “translate” each character of the first sentence to a list of candidate characters that could act as its counterpart in the second sentence. Then, filter out candidates that do not satisfy the linguistic constraints of couplets. Xingxing Z. et al.~\shortcite{Zhang:14} drew inspiration from sequence to sequence learning using the Recurrent Neural Networks(RNN) model that generated quatrains based on some given keywords. In contrast to the previous study by Zhou, M. et al. ~\shortcite{Zhou:09}, Xingxing Z. et al.~\shortcite{Zhang:14} did not make Markov assumptions about the dependency of the characters within a line.

In recent years, the development of an attention mechanism-based Encoder-Decoder framework in the NLP area has made it possible for computers to understand the text better. Google researchers Ashish V. et al. ~\shortcite{Vaswani:17} proposed a new simple network architecture called the Transformer. Although the Transformer is solely based on the attention mechanism, it perfectly preserved the recurrence and convolutions in previous neural networks models. J. Zhang, et al ~\shortcite{Zhang:18} introduced an open-source Chinese couplets generation system called VV-Couplets. It is an attention-based sequence to sequence neural model that maps the first line of couplets to the second line, generating similar output text giving the input. Compared to previous work on Chinese couplets, they addressed the entity names of person and location specially in their model. Although the Transformer model is widely used in the Natural Language Processing area, the existent framework is not fully explored for linguistic characteristics in Chinese. 

Our work is inspired by previous studies in Glyph ~\cite{Wu:19} that indicate since Chinese is a hieroglyph, glyph information will be helpful to extract character information. Wu et al.~\shortcite{Wu:19} regards Chinese characters as pictures and uses CNN to extract features. Wang et al. ~\shortcite{Wang:21} further improve the Transformer’s performance on Chinese couplets by intentionally adding POS taggings for special patterns in couplets, unregistered ancient words, and a polishing mechanism to improve the coherence. Sun et al. ~\shortcite{Sun:21} pretrained model by incorporating the glyph and pinyin information. Because Chinese characters have different meanings in different pronunciations, pinyin information is also crucial. Their findings can be instructive to improve the transformer mechanism.

Our approach departs from previous work is that we propose the Fusion embedding model, a couplet generation model with combining four features in the encoder. The four features extracted from Chinese couplets can effectively capture the semantic and grammatical rules of couplets. Furthermore, the fusion embedding mainly comprises the following parts: the Glyph embedding, the PinYin embedding, POS embedding and AnchiBERT embedding.

\section{Methods}

\subsection{Fusion Embeddings}

\begin{figure}[htp]
    \includegraphics[width=8cm]{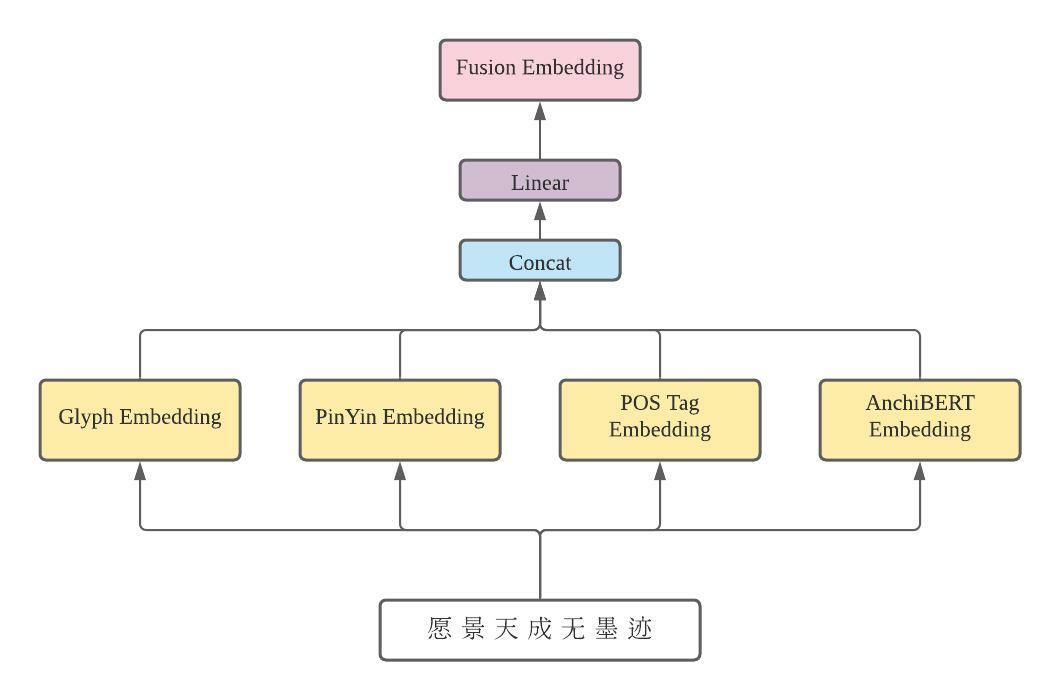}
    \caption{Fusion Embedding}
    \label{fig:fusion_emb}
\end{figure}

\begin{figure*}
  \centering
  \begin{subfigure}{0.4\linewidth}
    \includegraphics[width=1\linewidth]{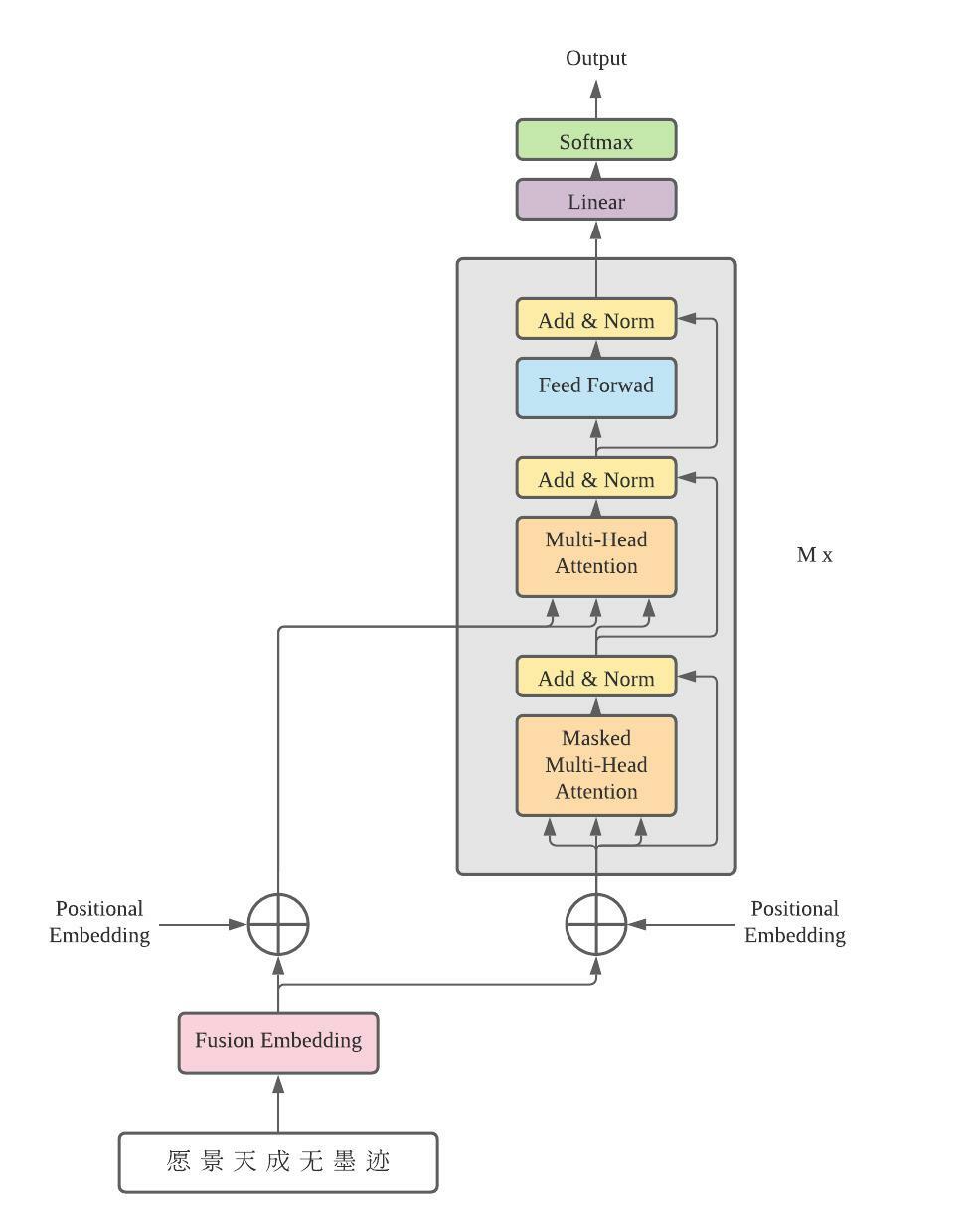}
    \caption{Fusion Decoder Model}
    \label{fig:short-a}
  \end{subfigure}
  \hfill
  \begin{subfigure}{0.4\linewidth}
    \includegraphics[width=1\linewidth]{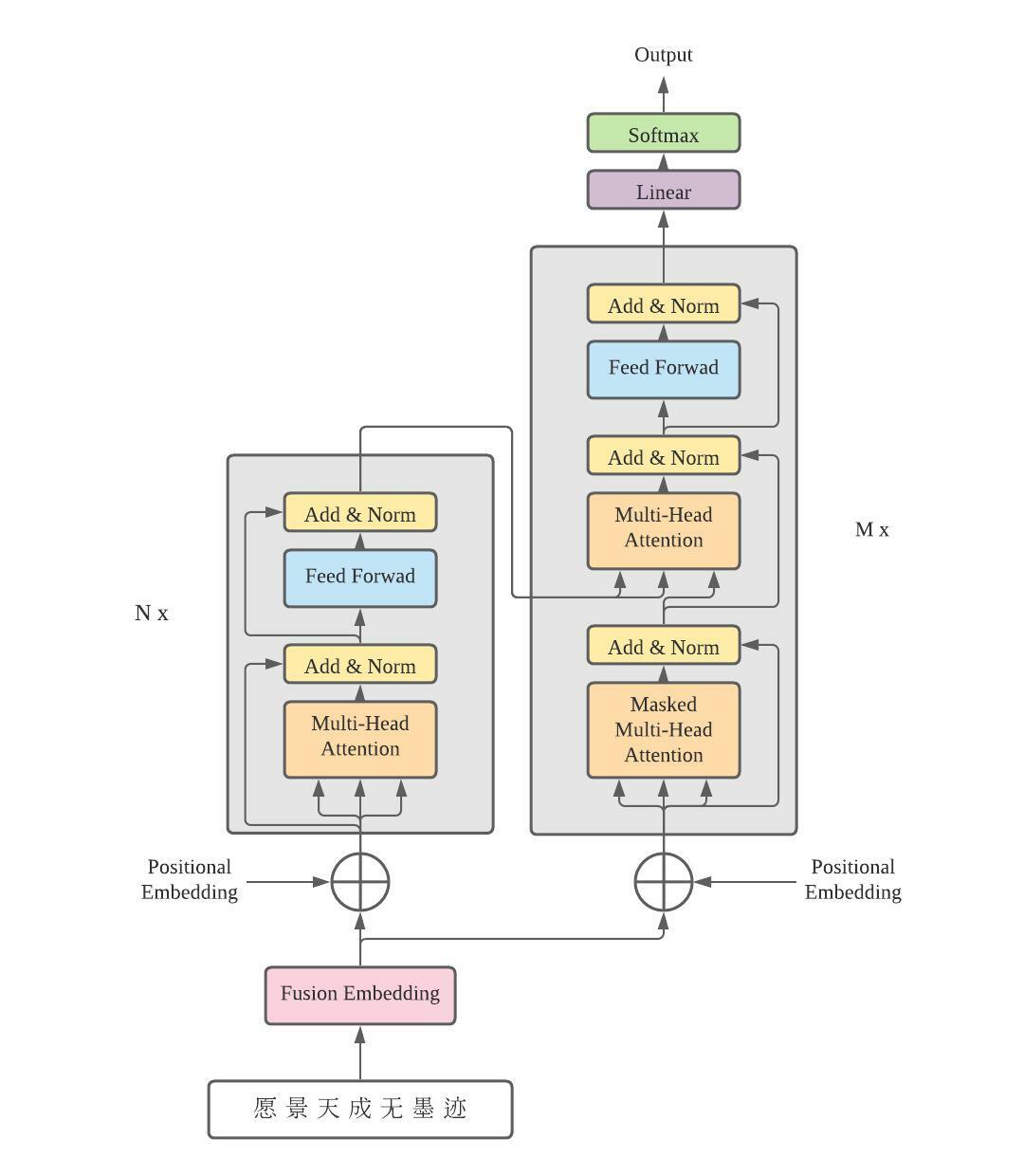}
    \caption{Fusion Transformer Model}
    \label{fig:short-b}
  \end{subfigure}
  \caption{Model Architecture}
  \label{fig:short}
\end{figure*}

In this paper, we constructed a custom embedding layer by combining 4 different embedding layers in order to capture the language understanding information and the couplet syntax. AnchiBERT~\shortcite{Tian:21} is a model used to process tasks of the ancient Chinese domain, such as generation, translation. We mainly built the fusion embedding based on the AnchiBERT model. Moreover, we evaluated the effect of Glyph, PinYin and the part-of-speech tagging (pos) embedding on the model performance. For the next three paragraphs, we cover the detailed information of Glyph Embedding, PinYin Embedding, Pos Tag Embedding, AnchiBERT Embedding, and Fusion Embedding.

\textbf{\emph{Glyph Embedding}} We first mapped all Chinese characters and punctuations into an embedding layer and initiated with customized weights. To obtain the customized weights, we use Chinese fonts, Fang-Song, to generate a black and white 24x24 image with floating point pixels ranging from 0 to 255. Then we obtained a 24 x 24 vector for each character.  In RGB representation, 0 represents the black and obtains the most important information about the character, but the linear layer is insensitive to zero. Therefore, different from Sun et al. (2021), which directly flattened the vector into 1x576 and fed to an FC layer, we normalized the vector value into a range from 0 to 1 and flipped the value (i.e., 255 to 0). Then we assigned the flattened vector as the initial weight for the corresponding character.

\textbf{\emph{Pinyin Embedding}} The rhythm is an essential component for Chinese couplets. A good couplet should not only obtain a match on semantic meaning between two sentences but also a match on the rhythm. Therefore, we embedded the feature Pinyin, a phonetic transcription for Mandarin, into our model. We used the open source pypinyin package to generate pinyin sequences. Pypinyin package \footnote{https://github.com/mozillazg/python-pinyin} utilized machine learning model and manually defined rules to infer the pinyin for each character from the context. Although we can use English characters to represent Chinese pinyin and 4 different symbols to represent the four different tones in Chinese, each character and symbol do not have any direct meaning between each other. Therefore, we enumerated all possible pinyin combinations in our training set, which is 1295, and put it into an embedding layer with 30 dimensions, where embeddings are uniformly sampled from range $[-\sqrt{\frac{3}{dim}} ,+\sqrt{\frac{3}{dim}}]$ where dim is the dimension of embeddings. 

\textbf{\emph{Pos Tag Embedding}} To be considered as a good couplet, the characters with relative same position from two consistent lines should be classified into the same category. Using Pos Tagging can help to alleviate the information gap (Wang et al., 2018). We use the open source jieba package \footnote{\url{https://github.com/fxsjy/jieba}} to generate pos tag sequences for each character from the context. Jieba package utilized a prefix with a directed acyclic graph for all possible word combinations and generative based model, like HMM to infer the pos tag information. There are 28 different pos tags in the jieba package. We put it into an embedding layer with 5 dimensions, where embeddings are uniformly sampled from range $[-\sqrt{\frac{3}{dim}} ,+\sqrt{\frac{3}{dim}}]$ where dim is the dimension of embeddings. 

\textbf{\emph{AnchiBERT Embedding}} AnchiBERT is trained on a large scale of ancient Chinese Corpora (Tian et al., 2021). It contains valuable contextual information regarding ancient Chinese Literature.  Therefore, we used the last hidden layer outputs from AnchiBERT as AnchiBERT Embedding.

\subsection{Architecture}

We proposed two different structures of models, as shown in Figure 3. Fusion Decoder Model consists only of the transformer decoder, while the Fusion Transformer Model employs the full Transformer model. With 6 layers in the encoder, 6 layers in the decoder, and 12 attention heads, the total number of parameters for the Fusion Decoder Model and Fusion Transformer Model are 115M and 60M respectively.

\subsection{Pretraining Embedding Layer}

Instead of training from scratch, we pre-compute the embedding weight of Glyph, PinYin and part-of-speech tagging (pos) layer by preprocessing the data to generate the embedding weight.
In order to capture the word vision information, we generate the word picture and use it as the pretrained Glyph embedding weight. Likewise, we pre-compute the PinYin and pos embedding weight to capture the syntax and grammetical information.

\section{Experiments}

\subsection{Approaches}

The models we introduced are mainly based on the Embedding and the Transformer. We built 4 models.
\begin{enumerate}[label=(\roman*)]
\item \emph{AnchiDecoder} AnchiDecoder is the model consisting of the AnchiBERT embedding and the Transformer Decoder.

\item \emph{AnchiTransformer} AnchiTransformer is the model consisting of the AnchiBERT embedding and the Transformer Encoder and Decoder. The model architecture is similar to the AnchiDecoder, but with the full Transformer model.

\item \emph{FusionDecoder} Different from the Anchi-models, the Fusion-models are built based on the fusion embedding we created using the Glyph, PinYin and pos tagging embedding. The FusionDecoder is the model consisting of the Fusion embedding and the Transformer Decoder.

\item \emph{FusionTransformer} Similar to the FusionDecoder model, FusionTransformer is the model consisting of the Fusion embedding and the Transformer Encoder and Decoder.

\end{enumerate}

We are able to compare the influence of the Decoder model and the Transformer model, and the effect of the Fusion Embedding we introduced.

\subsection{Dataset}

We perform experiments on the Chinese Couplets Dataset. 

The Chinese Couplets Dataset is available on \url{https://github.com/v-zich/couplet-clean-dataset}, which comes from an existing dataset in GitHub and contains around 740,000 couplets. Sensitive words in this dataset are deleted by searching the existing sensitive word vocabulary.

\subsection{Setting}

During training, we set the hidden dimension to be 9118 tokens to train the model with Adam optimizer, and the batch size is 128 with learning rate of 1e-4. We performed the training task mainly on Google Cloud Platform, with Nvidia Tesla T4. Our code is implemented based on Pytorch.

\subsection{Evaluation Metrics}

We propose to evaluate results from the following evaluation metrics.

\emph{Perplexity} Perplexity is based on the probability model and can measure the effectiveness of training results on the test set. According to the definition, Perplexity first calculates the inverse probability of the test set, and then normalizes it according to the number of words in the test set. Generally speaking, when the value of perplexity is lower, the training effect is better.

\emph{BLEU} Evaluation The Bilingual Evaluation Understudy is to calculate the similarity between two sentences, mostly used in machine translation. When using the BLEU, we calculate the similarity between the couplets generated by the model and the gold-standard descendants in the corpus to verify the accuracy of the model. The average of the scores generated by each couplet is the final accuracy of the entire training dataset. Moreover, we use unigram to get the accuracy of translation and n-gram to judge the fluency of sentences.

\section{Results and Discussion}
\begin{figure}[htp]
    \includegraphics[width=7cm]{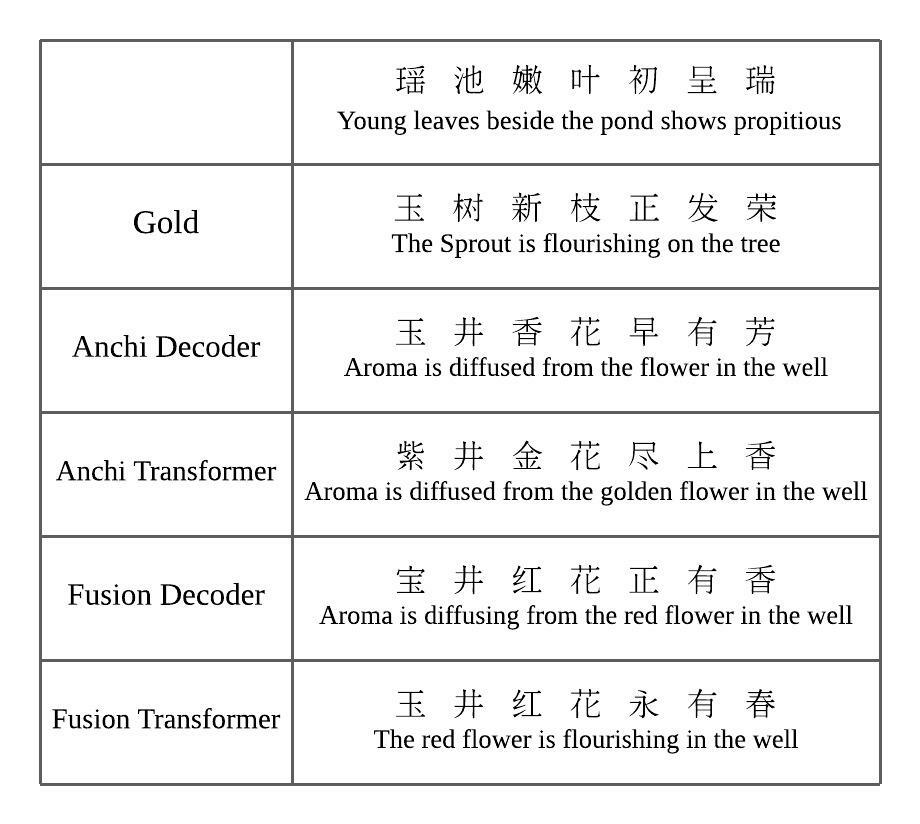}
    \caption{\small A sample result from the four models"}
    \label{fig:sample}
\end{figure}

Figure 4. gives an example to illustrates the characteristics of the 4 models. Gold is the  original answer from the dataset. The generated sentences are consistent in syntax and the grammatical rules of Chinese couplets. Also, the implication of all the results are cohesive to the topic of the given top couplet. 

In comparison, Table 1. shows the performance of each couplet generation models. We noticed that the BLEU score is low, that might be because of the complexity of Chinese language and the difference between the result and the gold from the original dataset. Based on the evaluation metrics, the fusion decoder model perform the best among the four systems.

\section{Conclusions}

\begin{table}[]
\normalsize
\begin{tabular}{|l|l|l|}
\hline
                  & BLEU & Bigram Perplexity  \\ \hline
Anchi Decoder     & 0.1030     & 485.09          \\ \hline
Anchi Transformer & 0.1012     & 496.25   \\ \hline
Fusion Decoder    & 0.1055     & 274.66    \\ \hline
Fusion Transformer    & 0.1031     & 282.18       \\ \hline
\end{tabular}
\caption{\label{tab: t1} Evaluation matrics}
\end{table}

The couplet generation is a challenging task because of the difficulty of Chinese disambiguation at the lexical and semantic level.  We proposed a new fusion decoder and encoder model to tackle this problem. Our fusion embedding combines 4 pretrained embedding layers to capture the positional, syntactic, and semantic information of each character. Given the first sentence, we produce the second sentence using sequence to sequence generation. 

Evaluating from the result we generated, we found the fusion decoder model perform the best and the result seems promising.

\end{document}